\documentclass[11pt,a4paper]{article}
\usepackage[hyperref]{acl2018} 
\usepackage{times}
\usepackage{latexsym}
\usepackage{amsfonts}
\usepackage{amsmath}
\usepackage{graphicx}
\usepackage{algorithm2e}
\usepackage{subfigure}
\usepackage{caption}
\usepackage{makecell}

\usepackage{url}

\aclfinalcopy % Uncomment this line for the final submission
\title{Learning How to Self-Learn: Enhancing Self-Training Using Neural Reinforcement Learning}

\author{Chenhua Chen \and Yue Zhang \\
        Singapore University of Technology and Design \\
        {\tt \{chenhua\_chen, yue\_zhang\}@sutd.edu.sg }}

\date{}

\begin{document}
\maketitle
\begin{abstract}
Self-training is a useful strategy for semi-supervised learning, leveraging raw texts for enhancing model performances. Traditional self-training methods depend on heuristics such as model confidence for instance selection, the manual adjustment of which can be expensive. To address these challenges, we propose a deep reinforcement learning method to learn the self-training strategy automatically. Based on neural network representation of sentences, our model  automatically learns an optimal policy for instance selection. Experimental results show that our approach outperforms the baseline solutions in terms of better tagging performances and  stability.
\end{abstract}

\section{Introduction}\label{section:intro}     
    
Scarcity of annotated data has motivated research into techniques capable
of exploiting unlabeled data, e.g., domain adaptation, semi-supervised and
unsupervised learning. Semi-supervised learning is useful for improving model performances when a target domain or language lacks of manual resources. 
Self-training is a commonly used strategy for various natural language processing (NLP) tasks, such as named entity recognition (NER) \cite{kozareva2005self}, part-of-speech (POS) tagging \cite{wang2007semi,qi2009semi} and parsing \cite{McClosky2006parsing,mcclosky2008self,Huang2009latentpcfg,Sagae2010parseradaptation}.

%It can potentially improve the baseline models which train using a set of existing labeled data, by    
%augmenting the training data with a set of raw data automatically labeled by the baseline model.    
The basic idea of self-training is to augment the original training set with a set of automatic predictions.  
In particular, traditional self-training approaches train baseline models using a few instances with gold labels. 
Then the baseline models are used to predict a set of unlabeled data.  
The automatically labeled instances are filtered based on certain strategies and then added to the training set for retraining new baseline models. 
This procedure continues until the number of automatically labeled instances reaches a budget or the model performance becomes stable.  Theoretically, it resembles expectation-maximization (EM), giving local minimum to the training objective.    

There have been different strategies for selecting automatically labeled data, the most typical one being the confidence values of the baseline models. How to define and measure the confidence of predictions is  crucial for a successful self-training approach. Traditional self-training solutions are designed manually based on task-specific heuristics \cite{rosenberg2005semi, Ardehaly2016domainadaptation,Jason2011emnlp,medlock2007weakly,daume2008cross}. This can lead to two drawbacks: 1) manual adjustment of instance selection strategy is costly; 2) for the best effect on an unknown dataset, the  source of information is limited to model confidence and a few other simple heuristics. As a result, linguistic characteristics of specific test sentences cannot be captured.
    
We aim to address this issue by leveraging neural models to represent test sentences, using reinforcement learning to automatically learn instance selection strategy. In particular, a self-training approach can be regarded as a decision process, where each step decides which automatically labeled instances can be selected. 
This decision process can be represented as a function that receives the automatically labeled instances as inputs and outputs a signal to indicate the acceptance or rejection of the prediction result.  
Since no gold labels exist for instance selection, we use  deep Q-network (DQN) \cite{nature14,guo2015generating,fang2017activelearning} to learn selection strategy automatically, on the basis of the performance improvements on a set of development data.A major advantage of our method as compared to traditional self-learning is that instance-level
characteristics can be combined with model level confidence information using a neural model that is adaptively trained using  reinforcement learning.

To evaluate the effectiveness of our proposal, we design and implement several self-training algorithms for the language-independent named entity recognition tasks and POS tasks. 
%We apply our solution to NER tagging in these scenarios. 
All the results show that our approach outperforms the baseline solutions in terms of performance and stability.  
Compared with manual heuristics, the deep Q-network model is capable of automatically learning the characteristics of training instances, and hence can potentially be more flexible in choosing the most useful data. 
Therefore, our approach is more general and can be applied to different NLP tasks. We release our code at http://anonymized. 

%you can remove \iffalse and \if to uncomment this
%\iffalse
The remainder of this paper is organized as follow:    
Section~\ref{sec:RW} reviews the state-of-the-art literature on
self-training and reinforcement learning for NLP tasks. Section~\ref{sec:slmethod} provides some background on the self-training process. Section~\ref{sec:dqnmethod}
introduces our self-training methodology, elaborating in detail how the deep
 Q-network for self-training is designed. 
Sections~\ref{sec:task}, ~\ref{sec:exps} and ~\ref{sec:casestudy} present our experimental setup for two typical NLP tasks, analyze the results, and explore the insights. 
Section~\ref{sec:conclusion} concludes the work and highlights some potential
research issues. 
%\fi

\section{Related Work}\label{sec:RW}

%Scarcity of annotated data has motivated the research into techniques capable
%of exploiting unlabeled data, e.g., domain adaptation, semi-supervised and
%unsupervised learning. 
%literature review logic
%1)one or two sentences to introduce the techniques
%2)How existing work addresses the issues in this areas
%3)The limitation of existing work
%4)The difference between existing work and ours
In this section, we briefly review previous work on
reinforcement learning, self-training and domain adaptation in the NLP area.
\textbf{Self Training} is a simple semi-supervised algorithm that has shown its effectiveness in parsing ~\cite{McClosky2006parsing,mcclosky2008self,Huang2009latentpcfg,Sagae2010parseradaptation},
part of speech tagging~\cite{wang2007semi,huang2009bigramhmm,qi2009semi}, named entity recognition~\cite{kozareva2005self,liu2013ner}, 
sentiment classification~\cite{van2016predicting,drury2011guidedself,liu2013reserved}, and other NLP tasks.

The performance of the self-training algorithms strongly depends on how automatically labeled data is selected at each iteration of the training procedure. Most
of the current self-training approaches set up a threshold and treat a set of
unlabeled examples as the high-confident prediction if its prediction is above
the pre-defined threshold value. Such a selection metric may not provide a
reliable selection~\cite{Chenmm2011nips}. 

Some researchers explore extra metrics as an auxiliary measurement to evaluate
instances from unlabeled data. For example, 
\newcite{Ardehaly2016domainadaptation} used coefficients learned from the model on the
source domain as a selection metric and report a positive effect when applying self training in the target for hierarchical multi-label classification task. 
 \newcite{Jason2011emnlp} proposed to produce a ranked list of n-best predicted parses 
 and selected the one yields the best external evaluation scored on the downstream external task (i.e., machine translation).
\newcite{rosenberg2005semi} examined a few selection metrics for object detection
task, and showed that detector-independent metric outperforms the more
intuitive confidence metric. Various pseudo-labeled example selection
strategies \cite{medlock2007weakly,daume2008cross} have been proposed.

\newcite{zhou2012self} explore a guided search algorithm to find informative unlabeled data subsets in the self-training process. The experimental results demonstrate that the proposed algorithm is in general more reliable and more effective than the standard self-training algorithm. These heuristic choices however require careful parameter tuning and domain specific information.

Most recently, \newcite{Levati2017tree} proposed an algorithm to
automatically identify an appropriate threshold from a candidate list for the
reliability of predictions. The automatic selected threshold is used in the next iteration of self-training procedure. They scored each candidate threshold by evaluating whether the mean of the out of bag error between the examples with reliability score greater than the considered threshold is
significantly different from the mean of the out of bag error of all the
examples.

Self-training techniques have also been explored for domain adaptation by some
researchers. \newcite{Ardehaly2016domainadaptation} used self-training process to adapt the label proportion model from the source
domain to the target domain.  \newcite{chattopadhyay2012multisource} generated 
pseudo-labels for the target data. \newcite{daume2008cross} proposed a self
training method in which several models are trained on the same dataset, and
only unlabeled instances that satisfy the cross task knowledge constraints are
used in the self training process.

\newcite{He:2011featureadaptation} derived a set of self-learned and domain-specific features that were
related to the distribution of the target classes for sentiment analysis task. 
Such self-learned features were then used to train another classifier 
by constraining the model’s predictions on unlabeled instances.

Note that although there have been many studies on how to apply self-training to
NLP tasks, their self-training strategies are designed manually based on the heuristics of the NLP tasks. This makes it difficult to apply these approaches to a different NLP task or a different application scenario. Differently, we propose to design a general self-training approach by learning the self-training
strategy automatically. In this way, our solution can be easily applied to
different application domains.

%{\bf Deep reinforcement learning for NLP} 
There has also been recent studies on reinforcement learning for natural language 
processing, such as machine translation \cite{grissom2014don,guo2015generating,gu2016learning,xia2016dual,bahdanau2016actor}, dialog systems \cite{li2016dialogue,su2016continuously}, sentence simplification \cite{zhang2017sentence,narayan2017split}, co-reference resolution \cite{clark2016deep} and latent structure induction \cite{Yogatama2016word2sentence,lei2016rationale}. 
In particular, \newcite{guo2015generating} proposed a deep Q-network model for text generation in sequence-to-sequence models. \newcite{fang2017activelearning} leveraged reinforcement learning for active learning, where a deep Q-network was used to learn the query policy for the active learners.  
%. For example, \newcite{su2016continuously} combined reinforcement learning with
%neural generation on tasks with real users, showing that reinforcement learning
%improves dialogue performance.\newcite{clark2016deep} applied reinforcement
%learning to directly optimize a neural mention-ranking model for co-reference
%evaluation metrics. xxx~\cite{} investigate knowledge transferring from source
%domain to target domain through.... 
%\newcite{Yogatama2016word2sentence} used reinforcement learning to induce implicit syntax structures via auxiliary tasks. \newcite{lei2016rationale} extracted rationales for sentiment prediction and question retrieval tasks. 
However, to the best of our knowledge, there has been no previous attempt to
embed self-training to deep reinforcement learning framework. 
The active learning work of \newcite{fang2017activelearning} is most similar to ours. However, different from active learning which needs to query human experts or oracle to provide gold labels,  
self-training associates instances with automatic prediction label. This makes our work substantially different from that of \newcite{fang2017activelearning}. 
%active learning (\cite{fang2017activelearning})

\section{Self-training}\label{sec:slmethod}
%formalize the problem as a self-learning problem
As explained in Section~\ref{section:intro}, we investigate how to learn a
self-training approach that can be applied to multiple scenarios where the gold labels are not available (or partially available) and the efforts for human to annotate the labels are expensive. Our goal is to retrieve useful information from the massive unlabeled dataset to improve the training performance.

In traditional self-training framework (Algorithm~\ref{alg:tsl}), a tagger is first initialized using a set of 
instances with gold labels. Then, this tagger is used to tag a set of unlabeled data, and the tagging confidence for each unlabeled
instance is evaluated. The automatically labeled instances  with the highest confidence is added to the
training set. The tagger is retrained using the updated training dataset and used to tag and select the unlabeled instances from the remaining dataset. This procedure is repeated until a given number of instances are selected and added to the training dataset. 

How to measure the confidence of a tagging is crucial to a self-training
approach. Traditional self-training approach usually requires human efforts to design specific confidence metrics based on the heuristics and application
scenarios. This, however, is challenging and the effectiveness of the solution may depend on many factors, which however is not well investigated yet. We propose to learn a self-training strategy automatically. 

%In traditional self-learning framework, the selection of an unlabeled instance
%can be regarded as a decision process, in the sense that the learning strategy
%decides whether an unlabeled instance is added to the training dataset or not in
%each step. Also, the selection of an unlabeled instance in current step depends
%on the selection of an unlabeled instance in the previous step, since the tagger is
%retrained once an unlabeled instance is selected and added to the training
%dataset. Therefore, a self-learning procedure can be regarded as a function that
%takes the current state of the self-learning framework as input and generates
%the output indicating whether an unlabeled instance is selected or not. In
%traditional self-learning approaches, this function is usually designed manually
%with some heuristics. In this paper, we propose to learn this function
%automatically using a deep reinforcement learning network. 

\begin{algorithm}[!t] 
\caption{Self-training}
\label{alg:tsl}
\KwIn{dataset train $T$, dev $D$, budget; unlabeled data $U$}
randomly shuffle $U$; $i \leftarrow 0$\;
\While{$i < budget$ and $U$ is not empty}{
    $tagger\leftarrow train(T, D)$\;
    $tag(tagger,U)$ \;
    $x \leftarrow select(U)| \max_{x}{\it confidence}(y)$\;
    $T \leftarrow T \cup (x, y)$\;
    $U \leftarrow U \backslash (x, y)$\;
    $i \leftarrow i + 1$\;
}
\KwOut{$tagger$;}
\end{algorithm}

\section{Learning how to self-train}\label{sec:dqnmethod}

%In a self-training framework, the selection of an unlabeled instance
%can be regarded as a decision process, in the sense that the learning strategy
%decides whether an unlabeled instance is added to the training dataset or not in
%each step. Therefore, a self-training procedure can be regarded as a function that
%takes the current state of the self-learning framework as input and generates
%the output indicating whether an unlabeled instance is selected or not. In
%traditional self-training approaches, this function is usually designed manually
%with some heuristics. In this section, we propose to learn this function
%automatically using a deep reinforcement learning neural network. 

In order to learn a self-training function automatically, we design a deep
reinforcement learning neural network to optimize the function parameters based on 
feedback from a set of development data.  
To formalize the self-training function easily, we
adopt a stream-based strategy. That is, the unlabeled training dataset is
regarded as a stream of instances. For every arriving unlabeled instance, the
self-training function generates an output indicating whether this instance is
selected or not. Given a stream of unlabeled dataset, our goal is to train the
function that can make the decision automatically with a good performance via
reinforcement learning.

\begin{figure*}[!t] 
\centering
\includegraphics[width=0.7\linewidth]{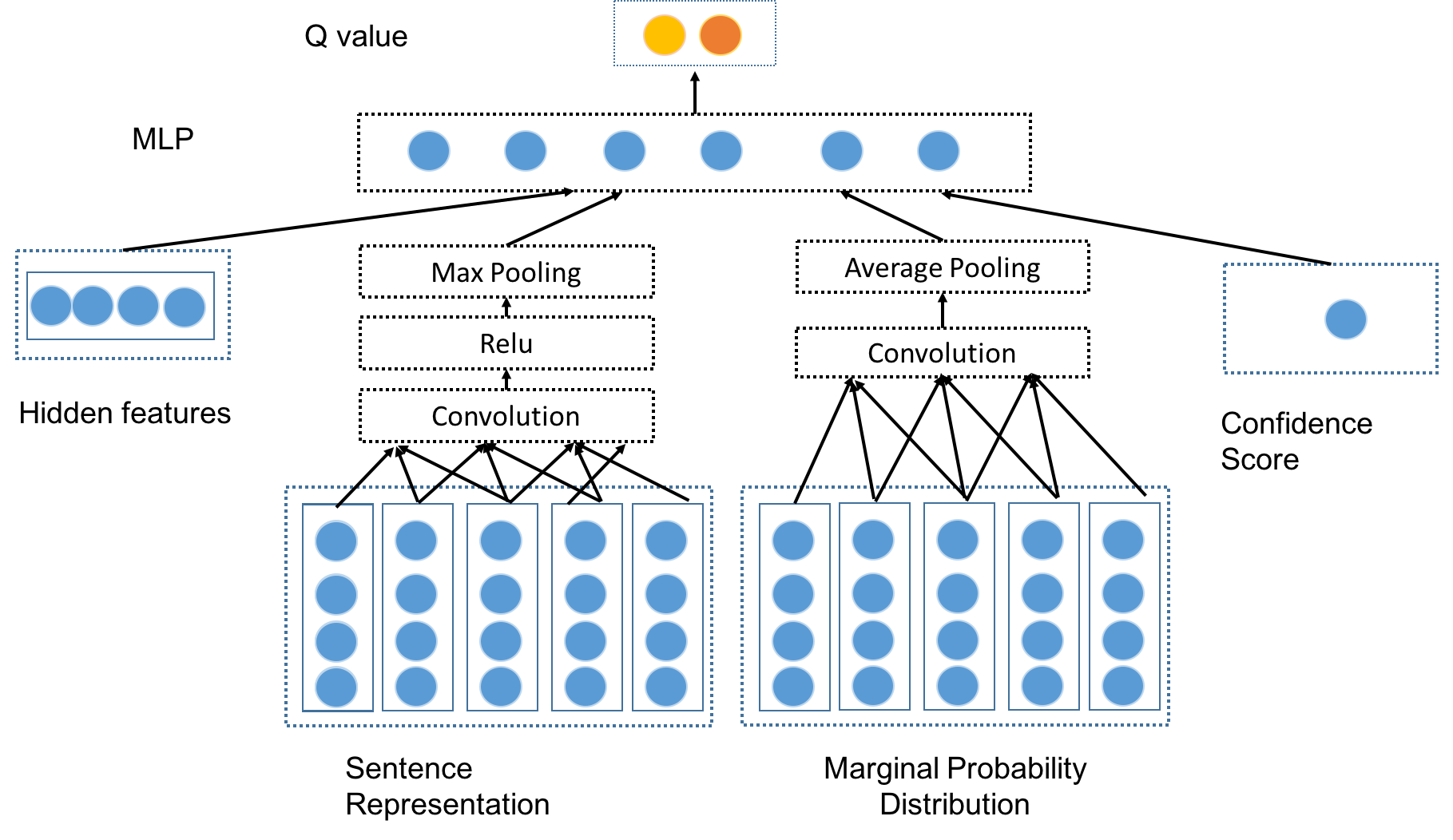}
\caption{The neural network for Q-function}
\label{fig:dqn}
\end{figure*}

%an introduction to reinforcement learning 
\subsection{DQN-based self-training}

\begin{algorithm}[!t] 
\caption{DQN-based self-training}
\label{alg:indomain}
\KwIn{dataset train T, dev set $D$, unlabeled data $U$,  budget;}
\KwOut{$tagger$;}
%\Begin{
    //train the DQN, c.f. Algorithm~\ref{alg:dqn}\\
    $DQN \leftarrow selftrain(U, D, T)$ \;
    $tagger \leftarrow train(T, D)$; $i \leftarrow 0$\;
    \While {$i<budget$ and $U$ is not empty} { 
        $x\leftarrow$ a random instance from $U$\;
        $U \leftarrow U \backslash {x} $\;
        $qvalue \leftarrow DQN(x, tagger)$\;
        \If {$argmax(qvalue) == 1$} {
             $y \leftarrow tag(tagger,x)$\;
             $T \leftarrow T \cup (x, y)$ \;
             $tagger\leftarrow train(T, D)$\;
             $i \leftarrow i+1$\;
        }    
    }
%}    
\end{algorithm}

Algorithm \ref{alg:indomain} shows the pseudocode for DQN-based self-training. Compared with Algorithm \ref{alg:tsl}, Algorithm \ref{alg:indomain} is different in three aspects. First, when making a choice, Algorithm \ref{alg:indomain} randomly accesses an unlabeled instance, while Algorithm \ref{alg:tsl} needs to visit a batch of instances. Second, Algorithm \ref{alg:tsl} directly uses model confidence to choose instances, while Algorithm \ref{alg:indomain} train a DQN to learn a Q-function ($qvalue$),  considering the confidence score and the test instance itself. Third, Algorithm \ref{alg:tsl} selects one instance for each iteration, while Algorithm \ref{alg:indomain} makes a choice only when the rejection score ($qvalue[0]$) is less than the acceptance score ($qvalue[1]$). 

\subsection{Model Structure}

We implement the
Q-function ($qvalue$ in Algorithm \ref{alg:indomain}) using a three-layered neural network, as shown in Fig.~\ref{fig:dqn}.

%state

The input layer represents the state of the learning framework. We adopt a
similar approach proposed by \newcite{fang2017activelearning} to define the state of the
framework, but we also integrate the hidden features from the tagger into the state. In particular, a state $s$ consists of four elements ($h_s$, $h_c$, $h_p$, $h_t$), where $h_s$ is the content representation of the arriving instance, $h_c$ is the
confidence of tagging the instance using the tagger, $h_p$ is the marginals
of the prediction for this instance, and $h_t$ is the hidden features from the tagger.

As shown in Fig.~\ref{fig:dqn}, to represent the content of an arriving
instance, a convolution neural network (CNN) with 384 filters are designed, with the convolution size being (3, 4, 5), and 128 filters 
for each size, respectively. Each filter uses a linear transformation with a
rectified linear unit activation function. The input to the CNN is the
concatenated word embedding of the instance via looking up a pre-trained word
embedding. These filters are used to extract the features of the instance
contents. The filter outputs are then merged using a max-pooling operation to
yield a hidden state $h_s$, a vector of size 384 that represents the
content of this instance.

The confidence of the tagger for an instance $h_c$ is defined based on the most
probable label sequence for this instance. In this paper, we adopt in the
experiment a Bi-LSTM-CRF tagger and its confidence value can be taken directly from the
CRF tagger. This value can be also defined and measured for different taggers. 

The marginals of the prediction for an instance is also calculated based on a
CNN. The labels of the arriving instance are predicted using the tagger and 20
filters with a size of 3 are designed to extract the features in the prediction.
These feature maps are then sub-sampled with mean pooling to capture the average
uncertainty in each filter. The final hidden layer $h_p$ is used to represent
the predictive marginals. 

In order to extract the hidden features from the tagger, we explore the last hidden layers from the neural network of the tagger and output them to represent the hidden features of the tagger.

%action
The four elements of the state $s=(h_s, h_c, h_p, h_t)$ are then used to calculate the second layer with a rectified linear unit activation function. This hidden layer
is designed as a vector of size 256. Then it is used to calculate expected
Q-value in the output layer, a vector of size 2, indicating whether the
instance should be selected or not. If an instance is selected, the predicted
labels of the sentences together with this sentence are added to the training
set to retrain the tagger.

\subsection{Training DQN}

{\bf Training Goal.} Reinforcement learning has been well adopted to learn a policy function in many
applications scenarios. In this paper, similar to the setting in
~\cite{fang2017activelearning}, we adopt the deep Q-learning approach~\cite{nature14} as our
learning framework (denoted as DQN). In this learning framework, the policy
function $\pi$ is defined via a Q-function: $Q^{\pi}(s,a)\to R$, where $s$ is
the current state of learning framework, $a$ is the action, and $R$ is the
reward of the framework when taking action $a$ from the current state $s$. The
training of the policy function $\pi$ is to find the parameters that can
maximize the reward of the each action in every state. This is done by
iteratively updating $Q^{\pi}(s,a)$ using the rewards obtained from the
sequence of actions during the training, based on the Bellman equation defined
as follows: 
\begin{equation} 
    Q^{\pi}(s,a) = \mathbb{E}[R_i|s_i=s, a_i=a, \pi]
\end{equation} 
where $R_i = \sum_{t=i}^T \gamma^{t-i} r_t$ is the discounted
future rewards and $\gamma\in[0,1]$ is the discounting rate. 

{\bf Reward.}  We define the
reward for each action as follow. If an instance is not selected, then the reward is set
to 0; otherwise, the reward of this action is defined as the performance
difference of the tagger after the instance is added to the training set. 

%budget for each episode (significance test) 

Following the approach of \newcite{nature14}, we train the DQN using an experience replay mechanism. The current state, its actions and corresponding rewards are recorded in a memory. The parameters of DQN are learned using stochastic
gradient descent to match the Q-values predicted by the DQN and the expected
Q-values from the Bellman equation, $r_i + \gamma max_aQ(s_{i+1}, a; \theta)$.
Samples are randomly selected from the experience memory to update the
parameters of the DQN by minimizing the loss function:
\begin{equation}
    \mathbb{L}(\theta) = (r+ max_{a'}Q(s',a')-Q(s,a))^2
\end{equation}    
The training procedure is repeated with  incoming instances. We conduct a
significance test to decide whether a current training  episode should be terminated.
In the significance test, we calculate the performance difference between a
series of consecutive actions. If the performance difference of the tagger is
not significant, we terminate the training episode and restart the training with
a new episode.

%may add an algorithm here
\begin{algorithm}[!t] 
\caption{Self-train a DQN}
\label{alg:dqn}
\KwIn{unlabeled data $U$, dev set $D$, initial set $init$;}
\KwOut{$DQN$;}
    $T \leftarrow init$\;
    \While {$episode < 10000$} { 
        $tagger\leftarrow train(T)$\;
        $score \leftarrow test(tagger, D)$\;
        
        \While {$significant(score)$}{
             $x\leftarrow$ a random instance from $U$\;
             $U \leftarrow U \backslash {x} $; $reward \leftarrow 0$\;
             $qvalue \leftarrow DQN(x, tagger)$\;
             \If{ $argmax(qvalue) == 1$} {
                 $y \leftarrow tag(tagger, x)$\;
                 $T \leftarrow T\cup (x, y)$ \;
                 $tagger \leftarrow train(T)$\;
                 $newscore \leftarrow test(tagger, D)$\;
                 $reward \leftarrow newscore - score$\;
                 $score\leftarrow newscore$\;
             }
             train $DQN(x, tagger, reward)$\;
        }
        $episode \leftarrow episode +1$ \;
    }   
\end{algorithm}

{\bf Algorithm.} Algorithm \ref{alg:dqn} shows the pseudo  code for DQN training. First, A initial set $init$ is used to train the baseline $tagger$. Then, a random initialized DQN evaluate the Q-value of an instance sampled from the unlabeled set $U$. Then when the corresponding Q-value of acceptance is larger than that of rejection, the instance is selected. 
 This raw instance together with its predicted labels is added to the training set. A new tagger is trained using the augmented training set. 
 When a selection happens, the performance difference between the new tagger and the old tagger on the development data set is calculated
as reward of the action. The reward is regarded as a feedback of the action and
is used to update the parameters of the DQN neural network . The DQN neural
network is self-trained for a given number of episodes. Once the performance on the development set is stable, the training of the
DQN stops.

\section{Tasks}\label{sec:task}
%\section{Tasks}
The deep reinforcement learning neural network for self-learning can be used for various application scenarios. In this section,
we illustrate two typical application scenarios. 
% application scenarios.

%\iffalse
% {\color{red}{\bf  \textbf{Self Training for Zero-shot Tagging} 

%In this scenario, the gold label of the training dataset is not available. Only
%a small development and test dataset is annotated with gold labels.  We need to apply a self-learning approach to train a tagger that can
%be used to tag the unlabeled instances with an acceptable performance.

%In order to learn the
%self-learning approach for this scenario, we randomly select a small number of
%instances with gold labels from the development dataset. This equals to setting the training set $T$ in Algorithm \ref{alg:indomain} to be a random subset $init$ from the develpoment set. 
%These instances are
%then used to initialize the tagger ($tagger \leftarrow train(T)$) for the self-training of the DQN (c.f., Algorithm~\ref{alg:dqn}). Once the training of the
%DQN is finished, we use it to select unlabeled instances from the training
%dataset and train the tagger based on those selected unlabeled instances
%(together with their predicted labels). /}}
%\fi

\textbf{Self Training for NER Tagging.} In this scenario, we have a training set with gold
labels, and a large number of unlabeled data. We need to apply a self-learning approach to improve the tagging performance using the unlabeled data. 

We train the baseline NER tagger using the training dataset with gold labels. In Algorithm \ref{alg:indomain}, the training set $T$ means the labeled training set here. 
%The purpose is to transfer the knowledge from the source domain to the target domain. 
%Then, this tagger is set as the initialized tagger to train the DQN
%using the unlabeled data from the target domain. If the DQN selects an unlabeled
%instance, the tagger is used to tag this instance. This instance and the
%predicted labels are then added to the training set to retrain the tagger.
%Similarly, the performance difference is calculated as the feedback of the
%action and used to update the parameters of the DQN neural network. 

\textbf{Self Training for POS Tagging.} In this scenario, we have a training set with gold
labels from the source domain, and a large number of unlabeled data from different target domains. We need to apply a self-learning approach to improve the tagging performance using the unlabeled data for different target domains. 

We train the baseline POS tagger using the source-domain dataset with gold labels. In Algorithm \ref{alg:indomain}, the training set $T$ means the source-domain training set here. 
%The purpose is to transfer the knowledge from the source domain to the target domain. 
%Then, this tagger is set as the initialized tagger to train the DQN
%using the unlabeled data from the target domain. If the DQN selects an unlabeled
%instance, the tagger is used to tag this instance. This instance and the
%predicted labels are then added to the training set to retrain the tagger.
%Similarly, the performance difference is calculated as the feedback of the
%action and used to update the parameters of the DQN neural network. 

In the above two tasks, the DQN neural network is self-trained for a given number of episodes (c.f., Algorithm~\ref{alg:dqn}). 
Once the training of the DQN is finished, we use the DQN and the baseline model to select unlabeled instances. 
These selected instances are added together with the training dataset to train the tagger to verify the tagging performance.

%use experiment 5 for the submission version
%\input{experiment5}
%use experiment 5.1 for the temp version
\section{Experiments}\label{sec:exps}

We conduct a series of experiments to evaluate our self-training based proposal for the aforementioned NER and POS tagging task. 

\subsection{Data}
In the NER experiments, we use the training, 
development and evaluation data sets from the CoNLL 2002/2003 shared tasks 
~\cite{tjong2002data,tjong2003data} for
four different languages: English, German, Spanish and Dutch. The data set for each language consists of newswire text annotated with four entity categories: Location (LOC), Miscellaneous (MISC), Organization (ORG) and Person (PER). We use the existing corpus portions, with train used for policy training, testb used as development set for computing rewards, and final results are reported on testa.

The europarl-v7 raw dataset~\cite{koehn2005europarl}  for machine translation is selected as the unlabeled data for the aforementioned four languages. 
A pre-trained embedding for each language~\cite{bojanowski2016enriching} is used in the experiments. 

%The pre-trained word vectors are  trained on Wikipedia using fastText.

In the cross-domain POS experiments, we use the SANCL2012 dataset. In this dataset, the ontonotes are used as the source domain, and the target domains include gweb-newsgroups, gweb-reviews, gweb-emails, gweb-weblogs.

\begin{figure*}[!t] 
    \centering
   \subfigure[English NER Tagging]{ 
      \includegraphics[width=0.45\linewidth]{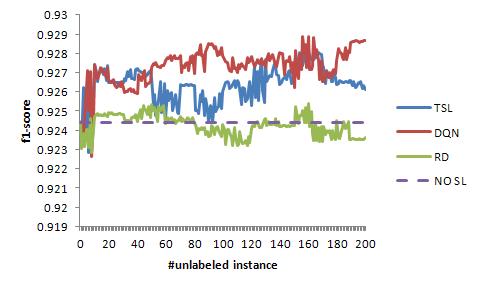}
      \label{fig:eng42}
   }
   \subfigure[Dutch NER Tagging]{ 
      \includegraphics[width=0.45\linewidth]{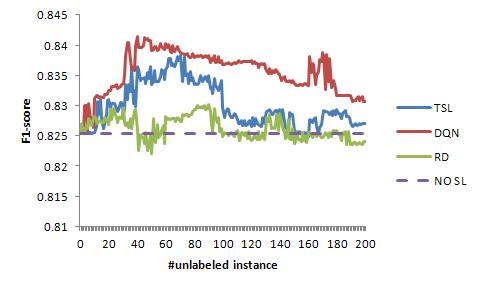}
      \label{fig:ned42}
   }
   \vspace{1mm}
   \subfigure[Spanish NER Tagging]{ 
      \includegraphics[width=0.45\linewidth]{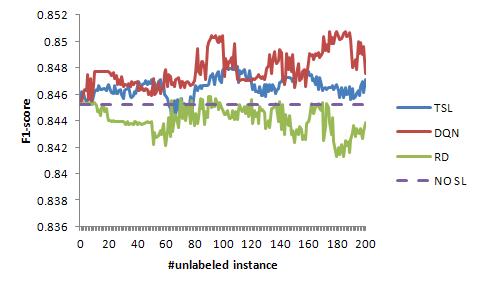}
      \label{fig:esp42}
   }
   \subfigure[German NER Tagging]{
      \includegraphics[width=0.45\linewidth]{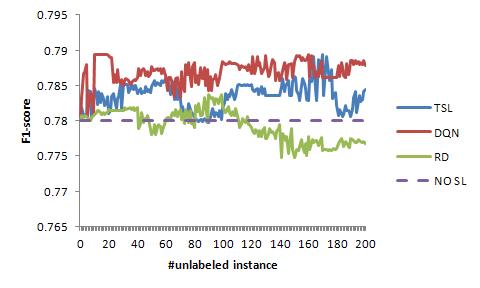}
      \label{fig:deu42} 
      %???
      %{\color{red}{\bf {NOT 200}}}
   }
%  \subfigure[Ontonotes to Gweb Adaptation]{ 
%      \includegraphics[width=0.45\linewidth]{gweb-answer.jpeg}
%      \label{fig:answer}
%   }   

\caption{Self-training for NER tagging}
\label{fig:exp2}
\end{figure*}

\subsection{Baselines, Evaluation Metrics and Training Settings}
To evaluate the effectiveness of the DQN-based solution, we also implement and test two baseline solutions, i.e., a random solution (denoted as $RD$) and a confidence-based self-training solution (denoted as $TSL$). In the $TSL$ solution, the confidence is calculated based on the most probable label sequence under tagger (i.e., NER or POS), i.e. given a unlabeled sentence $x_i$ with $n$ tokens, the confidence is defined as $\sqrt[n]{max_y{p(y|x_i)}}$. 

In both baseline approaches, the tagger (NER or POS) is initialed using the same instances as used in
the DQN-based approach. For the random solution, a stream of input sentences are randomly selected and tagged by the tagger (NER or POS). 

In the NER experiments, we use a state-of-the-art NER tagger, namely Sequence-Tagging NER tagger~\cite{lample2016neural}, as our baseline model. This NER tagger uses a Bi-LSTM on each sentence to generate contextual representation of each word, and then uses a CRF to decode the labels. We therefore consider extracting the contextual representation of words output by the Bi-LSTM as the hidden features of the NER tagger. That is, the DQN model uses the contextual representation of each word output by the Bi-LSTM in the NER tagger as $h_t$. These contextual representation of words together with the confidence and marginal probability of the tagging are furthered manipulated by the DQN model to predict the action. In the POS experiments, a CRF model is used to implement the POS tagger.     

%For DQN training, a cross-language pre-trained embedding \cite{Ammar2016MTLDS} is used both for in-domain and cross-domain tagging.  
In each training episode of DQN-based approach, if the latest 10 rewards in the episode is smaller than a given threshold (0.001), the performance change is regarded as not significant and this episode is terminated. To optimize the weights of the DQN, the DQN for each language is trained for 10000 episodes.

\subsection{Self-training for NER Tagging}

\textbf{Settings.} 
%We train the DQN combining the data from the source
%and target domains. 
%Similar to in-domain tagging, cross-lingual embeddings \cite{Ammar2016MTLDS} are used in the experiment.  
%Different from the zero-shot setting, 
We initialize the baseline NER tagger using the training set from the CoNLL2002/2003 dataset. 
%In this setting, the baseline model has a high prediction accuracy.  Similarly, 
We then self train a DQN for each of the four languages. Finally, we compare the performance of the DQN-based approach with the two baseline approaches.  
\fi
%In the
%Dutch domain, although the performance of all the three solutions drops with the
%increasing number of selected unlabeled sentences, the performance of our
%DQN-based solution drops more slowly that the two baseline solutions. Therefore,
%our DQN-based solution has a better strategy to select unlabeled sentences for
%the self-training in single-domain tagging tasks.
\begin{table}[!t]
\centering
\setlength{\tabcolsep}{3pt}
\begin{tabular}{|c|c|c|c|c|}
\hline
& {\bf{English}}    & {\bf Dutch}   & {\bf Spanish}   & {\bf German} \\ 
\hline
NO SL&  92.44 & 82.54  & 84.52   & 78.00 \\
\hline
RD   & 92.43 & 82.62  & 84.40   & 77.93  \\
\hline
TSL  & 92.64 & 83.01  & 84.65  & 78.35 \\ 
\hline
DQN  & {\bf 92.74} &{\bf 83.57 }  &  {\bf 84.80 }  & {\bf 78.68} \\ 
\hline
\end{tabular}
\caption{Performance comparison for NER tagging (\%)}
\label{table:exp2}
\end{table}

\begin{table*}[!h]
    \centering
    \begin{tabular}{|p{0.05\linewidth}|p{0.6\linewidth}|p{0.05\linewidth}|p{0.05\linewidth}|}
       \hline
        \bf ID & \thead{Instance} & \bf TSL & \bf DQN \\
        \hline 
        1  &  [{\color{blue}{Wayne Ferrei ra}}]
        {\color{blue}$_{\textsc{PER}}$} ([{\color{blue}{South Africa}}]{\color{blue}{$_{\textsc{LOC}}$}})  beat   [{\color{blue}{Jiri Novak}}] {\color{blue}{$_{\textsc{PER}}$}}
        ({\color{blue}{Czech$_{\textsc{LOC}}$}}).   &  Yes & Yes \\
        \hline
        2  &  The rapporteur wants assistants working in [{\color{blue}{Brussels}}]{\color{blue}{$_{\textsc{LOC}}$}} to be  covered by [{\color{violet}{Community rules}}]{\color{violet}$_{\textsc{{\color{violet}{ORG}}}}$}.     &  No  & Yes \\
        \hline
        3  &   says is first state to apply for new welfare.   &  Yes & No \\
        \hline
        4  &   Commissioner {\color{blue}{\text{[Frattini]$_\textsc{PER}$}}} wants {\color{blue}{\text{[Europe]$_\textsc{LOC}$}}} to attract a skilled workforce.    &  No & Yes \\
        \hline 
    \end{tabular}
    \caption{Selected instances for English NER task. {\color{blue}{Blue}} color and {\color{violet}{Violet}} color denote correct and wrong  predictions, respectively. }
    \label{tab:example}
\end{table*}

{\bf Results.} The results are shown in Table~\ref{table:exp2}. We can observe that our DQN-based solution has a
better performance than the two baseline solutions. In particular,  
Fig.~\ref{fig:eng42} shows the English NER tagging scenario, where the F1-score of the our DQN-solution increases about $0.6\%$ and then keeps stable at
about $92.8\%$. The F1-score of $TSL$ also increases, but the F1-score of the two baselines drop dramatically
after adding more unlabeled sentences. Note that although the $TSL$
approach has a sharp increase of performance at the moment of adding 50
unlabeled sentences, it is not stable compared to our DQN-based solution. We looked into the sentences selected by each solution, and found that $TSL$ selects almost the sentences with no entities (around 95.6\% of the selected sentences), because the sentences with no entities have a high confidence value. In contrast, $DQN$ favors the sentences with more and diverse entities, occurring in around 54.1\% of the selected sentences.

\begin{figure*}[!t] 
    \centering
   \subfigure[Newsgroup POS Tagging(testing)]{ 
      \includegraphics[width=0.45\linewidth]{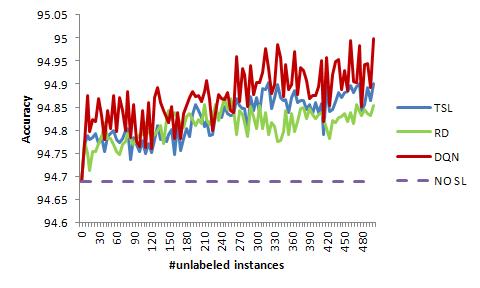}
      \label{fig:newsgroup-test}
   }
   \subfigure[Reviews POS Tagging(testing)]{ 
      \includegraphics[width=0.45\linewidth]{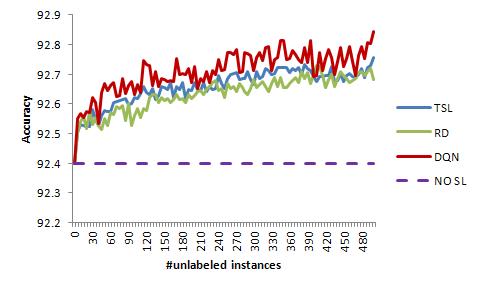}
      \label{fig:reviews-test}
   }
   \vspace{1mm}
   \subfigure[Weblogs POS Tagging(testing)]{ 
      \includegraphics[width=0.45\linewidth]{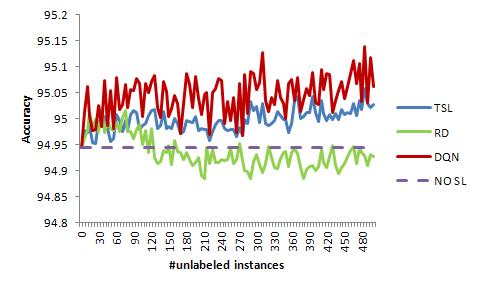}
      \label{fig:weblogs-test}
   }
   \subfigure[Emails POS Tagging(testing)]{
      \includegraphics[width=0.45\linewidth]{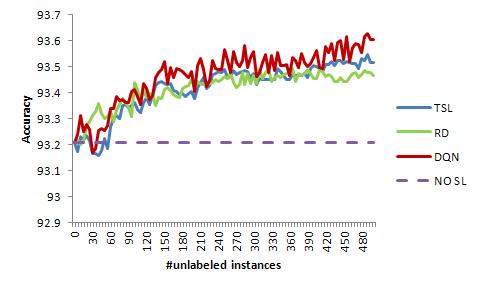}
      \label{fig:emails-test} 
   }
\caption{Self-training for POS tagging (testing)}
\label{fig:exp3}
\end{figure*}

\begin{figure*}[!t] 
    \centering
   \subfigure[Newsgroup POS Tagging(training)]{ 
      \includegraphics[width=0.45\linewidth]{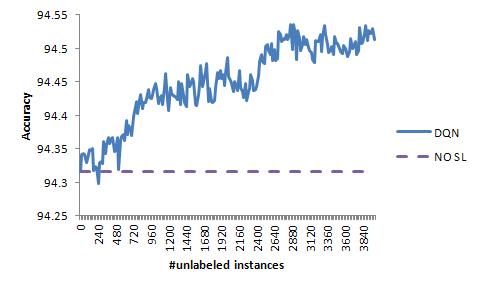}
      \label{fig:newsgroup-train}
   }
   \subfigure[Reviews POS Tagging (training)]{ 
      \includegraphics[width=0.45\linewidth]{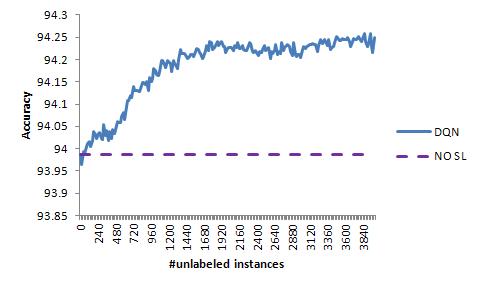}
      \label{fig:reviews-train}
   }
   \vspace{1mm}
   \subfigure[Weblogs POS Tagging(training)]{ 
      \includegraphics[width=0.45\linewidth]{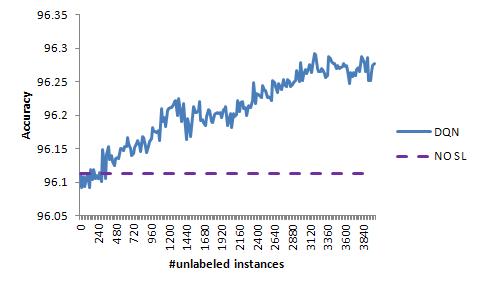}
      \label{fig:weblogs-train}
   }
   \subfigure[Emails POS Tagging(training)]{
      \includegraphics[width=0.45\linewidth]{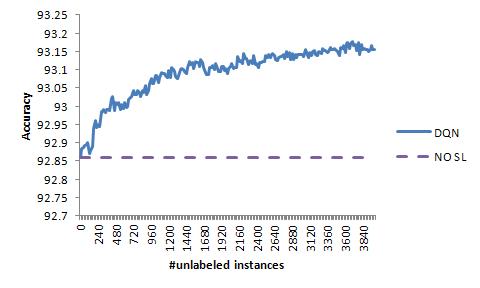}
      \label{fig:emails-train} 
    }
\caption{Self-training for POS tagging (training)}
\label{fig:exp4}
\end{figure*}

In the Dutch NER tagging scenario, as shown in Fig.~\ref{fig:ned42}, we can observe that all the three approaches have an improvement of the performance
when adding more and more unlabeled sentences. However, our DQN-based solution still outperforms the two baseline solutions because the performance of our solution increases more quickly and has a higher peak value than the two baseline solutions.

We can observe similar results for the Spanish and German NER tagging, as shown in Fig.~\ref{fig:esp42} and Fig.~\ref{fig:deu42}, respectively. In general, our DQN-based approach has a better tagging performance (about 0.3\% $\sim$ 1\% improvement on average) and is more stable than the other two baseline solutions. A complete comparison of the performance can be found in Table.~\ref{table:exp2}.

\subsection{Self-learning for POS tagging}

\textbf{Settings.} Similar to the setting in NER tagging, we also train and test the DQN combining the data for POS tagging. We initialize the POS tagger using the training set from the source domain (Ontonotes with gold labels). For each target domain (gweb-newsgroups, gweb-reviews, gweb-emails, gweb-weblogs), we train a DQN using 10000 episodes, in each episode 4000 sentences are selected. We compare the DQN-based self-learning solution with two baseline solutions (random and traditional self-learning) by adding the number of unlabeled sentences from 1 to 500.

\begin{table}[!t]
\centering
\setlength{\tabcolsep}{3pt}
\begin{tabular}{|c|c|c|c|c|}
\hline
& {\bf{Newsgroup}}    & {\bf Reviews}   & {\bf Weblogs}   & {\bf Emails} \\ 
\hline
NO SL&  94.69 & 92.40  & 94.94   & 93.20 \\
\hline
RD   & 94.81 & 92.63  &94.94   & 93.40  \\
\hline
TSL  & 94.83 & 92.65  & 95.00  & 93.41 \\ 
\hline
DQN  & {\bf 94.87} &{\bf 92.71 }  &  {\bf 95.04 }  & {\bf 93.47} \\ 
\hline
\end{tabular}
\caption{Performance comparison for POS tagging (\%)}
\label{table:exp3}
\end{table}

\textbf{Results.} The testing results are shown in Table~\ref{table:exp3} and Fig.~\ref{fig:exp3}. We can observe that our DQN-based solution has a better performance than the two baseline solutions in each target domain. In particular,  
Fig.~\ref{fig:newsgroup-test} shows that Ontonotes to Newgsgroup adaptation scenario, where the accuracy of the our DQN-solution increases about $0.03\%$ compared to NO SL solution (can reach 
about $94.99\%$ after adding 500 unlabeled instances). The accuracy of the two baselines also increase with the increasing number of unlabeled instance, but on average the accuracy of the DQN-solution is about $0.4\%$ and $0.6\%$ more than the $TSL$ and $RD$ solutions, respectively. Similar results can be observed for the Reviews and Emails scenarios, as shown in Fig.~\ref{fig:reviews-test} and Fig.~\ref{fig:reviews-test}. In the Weblogs scenario, as shown in Fig.~\ref{fig:weblogs-test}, the performance of $RD$ solution increases slightly after adding a few unlabeled instances, and then drops dramatically. In contrast, the performance of the $TSL$ and $DQN$ solutions increases slowly. On average, the $DQN$ solution has achieved an improvement of $0.04\%$ than the $TSL$ solution. 

In the POS experiments, we also recorded the performance of the tagger during the training. Fig.~\ref{fig:exp4} shows the training performance of the four scenarios. We can observe that the performance of the DQN-solution increases slowly by adding the increasing number of unlabeled instances in all four scenarios (up to $0.3\%$). This shows that the DQN-based self-learning is effective in improving the training performance.

%Then, we train
%the DQNs for three target domains (German, dutch, and Spanish) in a self-training
%way using the unlabeled dataset from each domain. Similar to the setting in the
%first experiment, when the DQN selects an unlabeled sentence in the
%target domain, we use the NER tagger trained from the source domain to predict
%its labels. This sentence together with its labels are added to the training
%set. The NER tagger and the DQN is then retrained using the updated training
%set. 
%For testing, once a DQN for a target domain is trained, we use it to select sentences from
%the target domain to train the NER model in self-training. Similarly, we
%initialize the NER tagger using the dataset from the source domain, then use the DQN to 
%select unlabeled sentences from the target domain. The two baseline approaches are similar to those in the
%first experiment. The difference is that the initialized NER tagger is trained
%from the source domain. The number of sentences
%selected from the target domain is the same for all the three approaches (200).

%\usepackage{color}
\section{Case Study}\label{sec:casestudy}

\begin{figure}[!t]
    \centering
    \includegraphics[width=\linewidth]{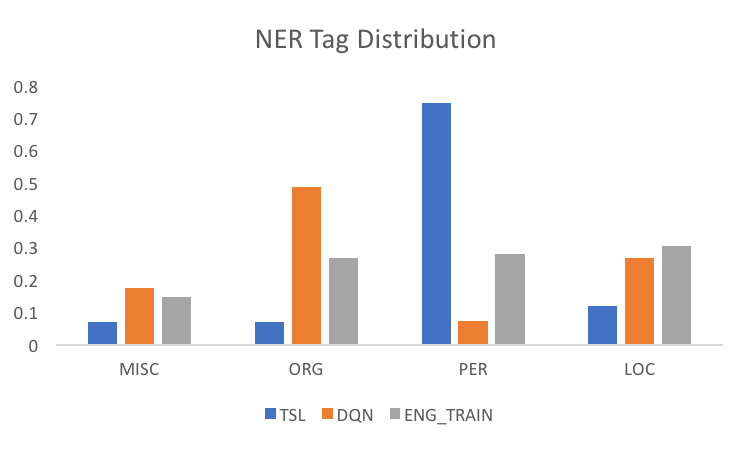}
    \caption{NER Tag probability distributions on instances selected by $TSL$, instances selected by $DQN$ and the original English training dataset.}
    \label{fig:tagdist}
\end{figure}

Table \ref{tab:example} shows some sample instances selected by $TSL$ and DQN for NER tagging. Two \textsc{PER} and two \textsc{LOC} entities in Instance 1 are tagged accurately.  Both $TSL$ and DQN select this instance. \textit{Community rules} is wrongly tagged as \textsc{ORG} in Instance 2, leading to a low confidence score. However, DQN successfully learns to accept this prediction result. The \textsc{LOC} entity in this instance is correctly tagged. A similar phenomenon can be found in Instance 4, both the \textsc{PER} and \textsc{LOC} entities are tagged with a very low confidence score, whereas DQN learns to accept this prediction result.  $TSL$ assigns Instance 3 a very high confidence score because there are no named entities in this instance at all, whereas DQN learns to reject this instance.   
%We further check the instances selected by $TSL$ and DQN. We find that $TSL$ favors the sentences with no entities, accounting for more than 90\% of the selected sentences. In contrast, $DQN$ favors the sentences with more and diverse entities, occurring in around 54.1\% of the selected sentences. 
In addition, we find that the entity tag distribution on the dataset produced by DQN is much closer to the distribution on the original English training dataset than that generated by $TSL$. Figure \ref{fig:tagdist} shows the overall distribution of the four entities on the three datasets. $TSL$ produces surprisingly large number of \textsc{PER} entities and small number of \text{MISC} and \textsc{ORG} entities, while $DQN$ generates less spiky distributions.

\section{Conclusion}\label{sec:conclusion}

%This paper investigates the use of self learning under reinforcement learning framework
%for cross lingual and domain adaptation tasks.

Learning a self-training strategy automatically can release the burden of human efforts in strategy design and is more flexible in choosing the most useful data. In this paper, we develop a deep reinforcement learning neural network to capture the characteristics of training instances automatically. Results show that our approach outperforms the baseline solutions in terms of a better tagging performance and stability. The current solution can be improved in several directions. For example, the input to the deep neural network does not include global tagging information of all the unlabeled instances. In addition, neural network structures other than CNN can be used to represent test sentences.

\bibliography{reference}
\bibliographystyle{acl_natbib}

\end{document}